\documentclass{article}

\usepackage{arxiv}
\usepackage[utf8]{inputenc}
\usepackage[T1]{fontenc}
\usepackage{hyperref}
\usepackage{url}
\usepackage{booktabs}
\usepackage{amsfonts}
\usepackage{nicefrac}
\usepackage{microtype}
\usepackage{graphicx}
\graphicspath{ {./images/} }
\usepackage{amsmath, amssymb}
\usepackage{lineno}
\usepackage[numbers]{natbib}
\usepackage{tikz}
\usetikzlibrary{arrows.meta,positioning,shapes.geometric}
\usepackage{calc}
\usepackage{xcolor}
\usepackage{adjustbox}
\usepackage{todonotes}
\usepackage{subcaption}
\usepackage{pifont}
\usepackage{threeparttable}
\usepackage{multirow}
\usepackage{makecell}
\usepackage{siunitx}
\usepackage{caption}
\usepackage{nameref}
\usepackage{longtable}
\usepackage{multirow}
\usepackage{booktabs}

\title{A Framework for Evaluating Predictive Models Using Synthetic Image Covariates and Longitudinal Data}

\author{
  Simon Deltadahl \\
  Department of Applied Mathematics and Theoretical Physics\\
  University of Cambridge\\
  Cambridge, UK \\
  \texttt{scfc3@cam.ac.uk} \\
  \And
  Andreu Vall \\
  Pumas-AI Inc \\
  \texttt{andreu@pumas.ai} \\
  \And
  Vijay Ivaturi\\
  Pumas-AI Inc \\
  \texttt{vijay@pumas.ai} \\
  \And
  Niklas Korsbo \\
  Pumas-AI Inc \\
  \texttt{niklas@pumas.ai} \\
}

\begin{document}
\maketitle

\begin{abstract}
In healthcare research, although abundant patient data is collected, privacy concerns limit its open availability and distribution for research purposes. We present a novel framework for synthesizing patient data consisting of complex covariates (e.g., the scan of a patient's eye) paired with longitudinal observations (e.g., the patient's visual acuity over time). This encompasses a wide set of tasks relevant to healthcare research. Our framework introduces a controlled degree of association already in the latent spaces that generate each data modality, naturally enabling the generation of complex covariates paired with longitudinal observations. This is crucial for the development and refinement of predictive models, and it provides a framework to generate openly available benchmarking datasets for healthcare research. We demonstrate our framework on optical coherence tomography (OCT) scans (an imaging method for the eye), but the framework is generally applicable to any domain. Using an open dataset of 109,309 2D OCT scan slices, we trained an image generative model combining a Variational Autoencoder and Stable Diffusion. A nonlinear mixed effect (NLME) model was chosen to simulate longitudinal observations from a low-dimensional space of random effects (a common approach in healthcare research). We generated OCT scan slices paired with longitudinal observations with controlled, decreasing levels of association. Specifically, we sampled 1.1M OCT scan slices associated to five sets of longitudinal observations, where the images, by design, could theoretically predict 100\%, 50\%, 10\%, 5.26\%, and 2\% of the between-subject variability of the longitudinal observations. We assessed the framework as follows: we modeled the synthetic longitudinal observations with yet another NLME model and computed the empirical Bayes estimates of the random effects (also a common approach in healthcare research); we trained a ResNet model to predict these estimates from synthetic OCT scan slices; we incorporated the ResNet predictions into the NLME model to make patient-individualized predictions of their longitudinal observations. The accuracy of these predictions on withheld data declined as intended as we reduced the degree of association between the images and the longitudinal measurements. Importantly, in all but the 2\% case, we reached within the 50\% of the theoretical best possible prediction on withheld data, indicating that we can find a signal even when it is weak. Data, models, and code are available at \url{https://doi.org/10.57967/hf/2089} and \url{https://github.com/Deltadahl/Image-Longitudinal}.
\end{abstract}

\vfill

\maketitle
\clearpage

\section{Introduction}\label{introduction}
The availability of data and computation resources has made recent advancements in deep learning possible~\cite{LeCun_Bengio_Hinton_2015}. Very large, open datasets have become common in domains such as computer vision~\cite{Deng_Dong_Socher_ImageNet, Lin_Maire_Belongie_Hays_Perona_Ramanan_Dollar_Zitnick_2014, LAION_2022}, audio processing~\cite{Mesaros_Heittola_Virtanen_2018, Rakotomamonjy_Gasso_2015}, and recommender systems~\cite{Koren_Bell_Volinsky_2009, Koenigstein_Dror_Koren_2011}, to cite a few. On the one hand, the data in these domains are rather noisy but abundant on the Internet. On the other hand, by nature, these data are arguably not as sensitive as, for example, medical patient data. The situation is indeed different in healthcare research. Despite vast amounts of medical patient data being generated, the data available to researchers remains scarce, mostly due to reasonable privacy concerns. Where possible, open datasets for healthcare research would be extremely valuable, both for researching specialized machine learning models and for establishing benchmarking mechanisms \cite{Johnson_Pollard_2016}.

In response to this, we introduce a novel framework to synthesize patient data consisting of complex covariates (e.g., the scan of a patient's eye) paired with longitudinal observations (e.g., the patient's visual acuity over time). This encompasses a wide range of tasks relevant to healthcare research. Here, we focus on generating medical imaging data, specifically, optical coherence tomography (OCT) scans (an imaging method for the eye), paired with synthetic longitudinal patient observations. The framework, however, is generally applicable to any domain. Our approach leverages the capabilities of Variational Autoencoders (VAEs)~\cite{kingma2013auto}, Stable Diffusion (SD) models~\cite{rombach2022high}, and Nonlinear Mixed-Effects (NLME) models~\cite{lindstrom1990nonlinear}. We trained a VAE on a publicly available dataset of 109,309 OCT images, licensed under CC BY 4.0 and available for research use~\cite{kermany2018large}. The dataset is anonymized, and it does not contain identifiable personal information. We also developed a novel approach to utilize the latent vectors $\boldsymbol{z}$ from the VAE as a condition for the diffusion process of the SD model. This allows us to generate more realistic medical images.

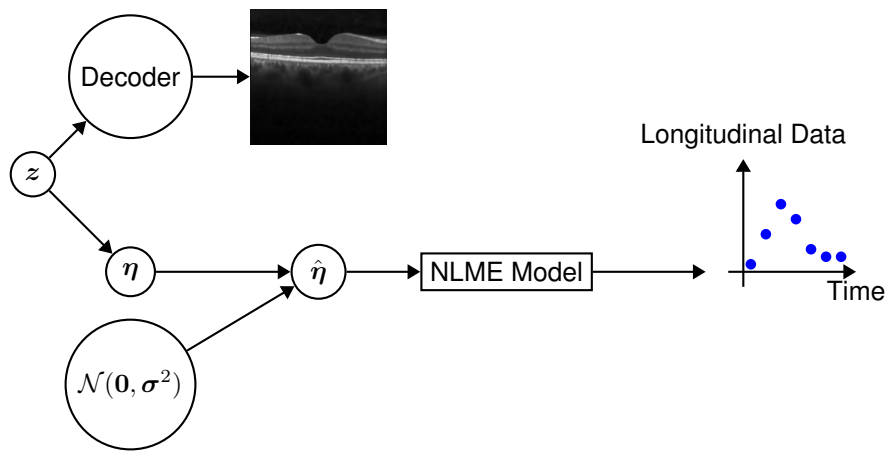
\begin{figure}[ht]
    \centering
    \begin{tikzpicture}[auto, thick, node distance=2.5cm, >=Triangle,font=\sffamily]

        \node[draw, circle] (z) {$\boldsymbol{z}$};

        \node[draw, circle, shift={(-0.47,-0.47)}, above right of=z] (decoder) {Decoder};

        \node[inner sep=0pt, right of=decoder] (reconstructedImage) {\includegraphics[width=1.8cm]{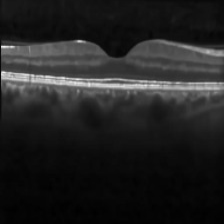}};

        \node[draw, circle, shift={(-0.47, 0.47)}, below right of=z] (eta) {\( \boldsymbol{\eta} \)};

        \node[draw, circle, below of=eta, node distance=1.5cm] (noise) {\( \mathcal{N}(\boldsymbol{0}, \boldsymbol{\sigma}^2) \)};

        \node[draw, circle, right of=eta, node distance=2.5cm] (etahat) {\( \hat{\boldsymbol{\eta}} \)};

        \node[draw, rectangle, right of=etahat, align=center, node distance=2.5cm] (nlme) {NLME Model};

        \begin{scope}[shift={(nlme.east)}, xshift=2cm]
            \draw[->, thick] (-0.2,0) -- (1.5,0) node[below] {Time};
            \draw[->, thick] (0,-0.2) -- (0,1.5) node[above] {Longitudinal Data};
            \foreach \x/\y in {0.1/0.1, 0.3/0.5, 0.5/0.9, 0.7/0.7, 0.9/0.3, 1.1/0.2, 1.3/0.2}{
                \fill[blue] (\x,\y) circle (2pt);
            }
        \end{scope}

        \draw[->] (z) -- (decoder);
        \draw[->] (decoder) -- (reconstructedImage);
        \draw[->] (z) -- (eta);
        \draw[->] (eta) -- (etahat);
        \draw[->] (noise) -- (etahat);
        \draw[->] (etahat) -- (nlme);
        \draw[->] (nlme.east) -- ++(1.5cm,0);

    \end{tikzpicture}
    \caption{The diagram illustrates the flow from the latent space vector $\boldsymbol{z}$, which is used both to generate medical images via the SD model and to derive a subset vector \(\boldsymbol{\eta} \subseteq \boldsymbol{z}\). Additional noise is applied to $\boldsymbol{\eta}$, resulting in $\hat{\boldsymbol{\eta}}$, which is then employed in the NLME model to produce synthetic longitudinal data. This process allows for controlled tuning of the correlation between the synthetic images and the generated longitudinal data outcomes.}
    \label{fig:z-to-img-and-long}
\end{figure}

Our framework leverages an existing association between the latent space of the VAE that generates OCT images and the latent space of the NLME model that generates longitudinal observations. The former typically has a much higher dimension than the latter. For example, in our experiments (Section~\ref{methods}), the VAE latent space is 128-dimensional, while the NLME latent space is only 3-dimensional. The procedure is as follows (Figure~\ref{fig:z-to-img-and-long}): we sample a vector $\boldsymbol{z}$ from a standard normal distribution with the dimension required for the VAE. On the one hand, this is used to condition the SD model to generate OCT images. On the other hand, $\boldsymbol{z}$ is indexed by a set of fixed dimensions $\left( i_1, \dots, i_k \right)$ to obtain a vector $\boldsymbol{\eta} = \left( z_{i_1}, \dots, z_{i_k} \right)$ that is used as the random effects in the NLME model to synthesize longitudinal patient data. This design ensures an association between each generated OCT image and its corresponding synthesized longitudinal observations. To adjust the extent to which the longitudinal observations can be predicted from the images, we introduce additional noise to $\boldsymbol{\eta}$, creating $\hat{\boldsymbol{\eta}}$. This adjustment allows us to control the predictive variability and subsequently generate synthetic longitudinal data that reflect different levels of image-based predictive power.

The ability to generate medical images associated with synthetic patient trajectories provides a valuable tool for testing and empirical evaluation of predictive models. By utilizing synthetic data, researchers can gain insights into the potential of different model classes to address specific tasks without initially relying on real patient data. Our framework enables the empirical assessment of a model's feasibility and informs further model development stages. It is important to recognize that synthetic data offer a rough indication of a model class's capability, facilitating model refinement before rigorous validation and benchmarking on real datasets.

The primary contributions of our study are:

\begin{itemize}
\item We introduce a framework combining VAEs, SD models, and NLME models to generate complex patient covariates paired with patient longitudinal data.
\item We demonstrate our framework by generating medical images and longitudinal patient data.
\item We propose a novel method for conditioning SD models directly with VAE latent space vectors (no text required, as is usually the case in SD).
\item Our methodology allows explicit control over the degree of association between images and longitudinal patient observations. This lets us adjust the complexity of the generated benchmark datasets.
\item We release a public dataset of 1.1 million synthetic images, each paired with five sets of longitudinal data at different noise levels, resulting in a total of 5.5 million longitudinal data pairs, facilitating widespread testing, benchmarking, and advancement in predictive healthcare modeling.
\end{itemize}

\section{Methods} \label{methods}

\subsection{Variational Autoencoder (VAE)}

We trained a custom VAE, utilizing a ResNet18 encoder, a 128-dimensional latent space, and a custom decoder. The complete architecture is detailed in the Appendix. The dataset creators' train/test split for the OCT images was maintained throughout our experiments \cite{kermany2018large}.

We preprocessed the images by resizing them to $224\times224$ pixels. We augmented the training dataset by applying horizontal flipping and random vertical cropping to remove 34 rows, followed by resizing the images back to $224\times224$ pixels. For the validation set, we used central vertical cropping instead of random cropping.

Our $\beta$-VAE \cite{Higgins_Matthey_2016}, employed a loss consisting of the standard Kullback-Leibler (KL) divergence loss with, and a hybrid reconstruction loss function. The hybrid loss combined two perceptual losses \cite{Pihlgren_Nikolaidou_Chhipa_Abid_Saini_Sandin_Liwicki_2023} derived from separate layers of a pre-trained VGG16 network on ImageNet \cite{vgg16, Deng_Dong_Socher_ImageNet}, along with a MSE loss

\begin{equation}
\begin{aligned}
\mathbb{E}_{q_{\phi}(\boldsymbol{z}|\boldsymbol{x})}[\log p_{\theta}(\boldsymbol{x}|\boldsymbol{z})] &= \alpha \mathcal{L}_{\text{MSE}}(\boldsymbol{x},\boldsymbol{\hat{x}}) \\
&+ \gamma \mathcal{L}_{\text{perceptual}}(g_2(\boldsymbol{x}),g_2(\boldsymbol{\hat{x}})) \\
&+ \tau \mathcal{L}_{\text{perceptual}}(g_7(\boldsymbol{x}),g_7(\boldsymbol{\hat{x}}))\\
&+C,
\end{aligned}
\label{eq:ELBO_reconstruction_components}
\end{equation}

where $g_j$ is the output of the $j^{\text{th}}$ ReLU layer of the VGG16 network, and $C$ is a constant. Initially, the weighting factors $\alpha$, $\gamma$, and $\tau$ were set to $\frac{1}{3}$. To scale the losses appropriately, we adjusted $\alpha$ to $\frac{400}{3}$ and $\gamma$ to $\frac{25}{12}$. These modifications not only brought the losses to a comparable scale but also visibly improved the quality of the reconstructed images compared to those generated without these weightings.

The VAE was trained on an NVIDIA RTX 3090 GPU with a batch size of five OCT scans. We used the Adam optimizer with a learning rate of 0.001 and the moment coefficients suggested in the paper \cite{Kingma_Ba_2017}.

The training process consisted of two phases. In the first phase, the $\beta$ value from the $\beta$-VAE approach was linearly increased from zero to its maximum of $\beta = 0.0376$, to establish a stable training environment \cite{Burgess_Higgins_2018}. Following the initial training period, the model underwent an additional 100 epochs.

\subsection{Stable Diffusion (SD)}
To enhance the quality of the generated images, we utilized Stable Diffusion 1.5 \cite{rombach2022high}. The original OCT images, with dimensions of $496\times512$ pixels, were resized to $512\times512$ for training the SD model. Instead of using the standard text tokenizer and text encoder in the SD model, we designed a novel approach by directly feeding the latent space vectors from the VAE as conditioning information. However, since the output of the text encoder in the SD model is of size $77\times768$ and our latent space is of size 128, we adapted our latent space vectors to match the expected dimensions. This was achieved by copying each latent space vector to all 77 rows and padding them with zeros to reach a dimension of 768. This modification allowed for a direct integration of the VAE's learned representations into the SD model, replacing the traditional text-based conditioning.

Fine-tuning of the SD model was performed using the OCT images with corresponding latent space vectors as conditioning. The training was conducted on an NVIDIA RTX A5500 GPU with a batch size of 4, utilizing the AdamW optimizer with original parameters and a learning rate of $10^{-5}$ \cite{Kingma_Ba_2017}. An exponentially moving average (EMA) with a decay rate of 0.9999 was employed during the training process \cite{Loshchilov_Hutter_2019}, which lasted for 116,000 steps.

\subsection{Nonlinear Mixed-Effects (NLME) models}
NLME models are widely used in fields such as pharmacokinetics, epidemiology, and medical research to analyze longitudinal data and account for individual variability within a population \cite{pinheiro2006mixed}. These models are particularly useful for complex hierarchical data structures where measurements are taken repeatedly over time on the same individuals.

An NLME model comprises two primary components: the structural (or deterministic) model and the statistical (or stochastic) model. The structural model is described by fixed effects, while the statistical model accounts for random effects, representing individual variability, and residual errors, representing observational noise.

Consider \( Y_{ij} \) as the \( j^{\text{th}} \) response from the \( i^{\text{th}} \) individual. The NLME model can be represented as

\begin{equation}
Y_{ij} = f(\boldsymbol{X}_{ij}, \boldsymbol{\beta}, \boldsymbol{\eta}_{i}) + \varepsilon_{ij},
\label{NLME_model}
\end{equation}

where \( f \) denotes a potentially nonlinear function, \( \boldsymbol{X}_{ij} \) is a vector of covariates (features), \( \boldsymbol{\beta} \) is a vector of fixed effects parameters, \( \boldsymbol{\eta}_{i} \) represents the random effects parameters (latent variables) specific to the \( i^{\text{th}} \) individual, and \( \varepsilon_{ij} \) signifies the residual error.

The random effects, $\boldsymbol{\eta}_{i}$, are assumed to follow a multivariate normal distribution with zero mean $\boldsymbol{\eta}_{i} \sim \mathcal{N}(\boldsymbol{0}, \boldsymbol{\Omega})$, where $\boldsymbol{\Omega}$ is the variance-covariance matrix. The residual errors, $\varepsilon_{ij}$, are typically normally distributed with zero mean and variance $\sigma^{2}_\varepsilon$, expressed as $\varepsilon_{ij} \sim \mathcal{N}(0, \sigma^{2}_\varepsilon)$ \cite{Davidian_Giltinan_2003, Demidenko_2013, pinheiro2006mixed}.

Parameter estimation in NLME models often uses Maximum Likelihood Estimation \cite{Davidian_Giltinan_2003}. However, due to the complexity introduced by the random effects, the likelihood function involves a multi-dimensional integral that is generally not analytically solvable \cite{Pinheiro_Bates_1995}. To address this, the marginal likelihood is employed, integrating out the random effects

\begin{equation}
L(D | \boldsymbol{\beta}, \boldsymbol{\Omega}, \sigma^{2}_\varepsilon) = \int L(D | \boldsymbol{\beta}, \boldsymbol{\eta}_{i}, \sigma^{2}_\varepsilon) p(\boldsymbol{\eta}_{i} | \boldsymbol{\Omega}) d\boldsymbol{\eta}_{i},
\end{equation}

where $L(D | \boldsymbol{\beta}, \boldsymbol{\eta}_{i}, \sigma^{2}_\varepsilon)$ is the conditional likelihood, $p(\boldsymbol{\eta}_{i} | \boldsymbol{\Omega})$ is the distribution of the random effects, and $D$ represents the data \cite{pinheiro2006mixed}.

The dynamics of the NLME model are captured by two ordinary differential equations

\begin{align}
\frac{\mathrm{d} D_{i}}{\mathrm{d}t} &= -Ka_i \times D_{i}, \label{ODE1}\\
\frac{\mathrm{d} C_{i}}{\mathrm{d}t} &= Ka_i \times D_{i} - \frac{I_{\text{max},i} \times C_{i}}{IC_{50,i} + C_{i}}, \label{ODE2}
\end{align}
where
\begin{align}
Ka_i &= e^{\eta_{i,1}}, \\
I_{\text{max}, i} &= 2.1 e^{\eta_{i,2}}, \\
IC_{50, i} &= 0.4 e^{\eta_{i,3}}.
\end{align}

The parameters $\eta_{i,1}, \eta_{i,2},$ and $\eta_{i,3}$ are derived from the latent space of the VAE, signifying the random effects in our model. To simulate the longitudinal data, we initialize $C_i(0)=0$ and $D_{i}(0)=1$. For each individual $i$ and each observation $j$, the model predicts an output variable

\begin{equation}
Y_{ij} \sim \mathcal{N}(C_{ij}, \sigma^2_\varepsilon), \label{eq:Yij}
\end{equation}

where $\sigma_\varepsilon$ denotes the standard deviation of the observational error and is set to a value of $0.01$.

\subsection{Leveraging VAE Latent Space for NLME Modeling}
The latent space of the VAE can be viewed as random effects in the NLME model. This connection allows us to use the same latent vectors to generate both synthetic images via the SD model and corresponding longitudinal data via the NLME model, as illustrated in Figure \ref{fig:z-to-img-and-long}. This methodology ensures a consistent relationship between images and longitudinal observations, facilitating the creation of large-scale synthetic datasets with known, adjustable correlations between the two data types.

Since the VAE latent space is larger than the number of random effects, we opted to select the latent dimensions that were most influential as our random effects. The identification of these influential dimensions was performed using two methods:

\begin{itemize}
    \item \emph{Method 1}: 1,000,000 latent vectors, denoted as $\boldsymbol{z}$, were drawn from a standard Gaussian distribution. These vectors were processed through the VAE decoder and subsequently the encoder, resulting in a set of vectors labelled as $\boldsymbol{\hat{z}}$. The MSE between $\boldsymbol{z}$ and $\boldsymbol{\hat{z}}$ was then computed. The dimensions with the smallest MSE were considered the most influential.

    \item \emph{Method 2}: 10,000 new latent vectors $\boldsymbol{z}$ were generated. For each of these vectors, an iterative process was applied where one out of the 128 latent variables was altered while keeping the remaining 127 constant. This modification involved assigning a random value from a normal distribution, yielding the vector $\boldsymbol{\hat{z}}$. The original $\boldsymbol{z}$ and the modified $\boldsymbol{\hat{z}}$ were both passed through the decoder, and their MSE was calculated. The dimensions with the largest MSE were regarded as the most influential.
\end{itemize}

Upon comparing the most influential features derived from both methods, a consistent result was observed. The same top three dimensions emerged from both approaches and were thus selected to represent our random effects.

The synthetic data was generated by sampling $\boldsymbol{z}$ from a standard normal distribution, $\boldsymbol{z} \sim \mathcal{N}(\boldsymbol{0}, \boldsymbol{1})$. The SD model was then conditioned on $\boldsymbol{z}$ to produce synthetic images. Concurrently, the selected subset of the latent variables was assigned as random effects $\boldsymbol{\eta}$.

\subsection{Introducing Variable Noise to Random Effects}
In reality, it is unlikely that all information in the longitudinal data can be explained solely by an image. To account for this, we introduced different levels of Gaussian noise to $\boldsymbol{\eta}$. Each image was connected to five sets of random effects, each with a different noise level. The noise was introduced by transforming the random effects according to

\begin{equation}
\boldsymbol{\hat{\eta}} = \frac{\boldsymbol{\eta} + \boldsymbol{r}}{\sqrt{1+\sigma^2}}, \label{eq:eta_hat}
\end{equation}

where $\boldsymbol{r}\sim \mathcal{N}(\boldsymbol{0},\boldsymbol{\sigma^2})$ and $\boldsymbol{\eta}\sim \mathcal{N}(\boldsymbol{0},\boldsymbol{1})$, resulting in $\boldsymbol{\hat{\eta}}\sim \mathcal{N}(\boldsymbol{0},\boldsymbol{1})$. We used five different values for $\sigma^2$: 0, 1, 9, 18, and 49. By applying these noise levels, we generated five sets of longitudinal data for each image, each with a different degree of correlation between the image and the longitudinal observations. This approach allows for a more realistic representation of the relationship between images and longitudinal data, as well as enables the evaluation of predictive models under various noise conditions. The final dataset comprised 1,100,000 synthetic images, each paired with five sets of synthetic longitudinal data corresponding to the different noise levels.

\subsection{Evaluate connection between images and longitudinal data}
To evaluate the connection between the synthetic images and synthetic longitudinal data, we divided the dataset into three subsets: 950,000 samples for training, 50,000 for validation, and 100,000 for testing. Prior to this, we have viewed the NLME model as a decoder that translates random effects into longitudinal data. However, we now also leverage its capacity as an encoder, using empirical Bayes estimates to convert longitudinal data back into approximated random effects. By utilizing this dual functionality, we employed the NLME model to approximate the random effects as  \(\boldsymbol{\eta_{\text{approx}}}\)  from the longitudinal data, which then served as our training objectives.

\begin{figure}
\centering
\begin{tikzpicture}[auto, thick, >=Triangle,font=\sffamily]

    \node[inner sep=0pt] (longitudinalData) {
        \begin{tikzpicture}
            \draw[->, thick] (-0.2,0) -- (1.5,0) node[below=0.2cm] {Time};
            \draw[->, thick] (0,-0.2) -- (0,1.5) node[above] {Longitudinal Data};
            \foreach \x/\y in {0.1/0.1, 0.3/0.5, 0.5/0.9, 0.7/0.7, 0.9/0.3, 1.1/0.2, 1.3/0.2}{
                \fill[blue] (\x,\y) circle (2pt);
            }
        \end{tikzpicture}
    };

    \node[draw, circle, right=2.1cm of longitudinalData] (foce) {NLME};
    \draw[->] (longitudinalData.east) -- (foce.west);

    \node[draw, circle, right=2.1cm of foce] (eta_approx) {\( \boldsymbol{\eta}_{\text{approx}} \)};
    \draw[->] (foce.east) -- (eta_approx.west);

    \node[inner sep=0pt, below=0.8cm of longitudinalData] (initialImage) {\includegraphics[width=1.8cm]{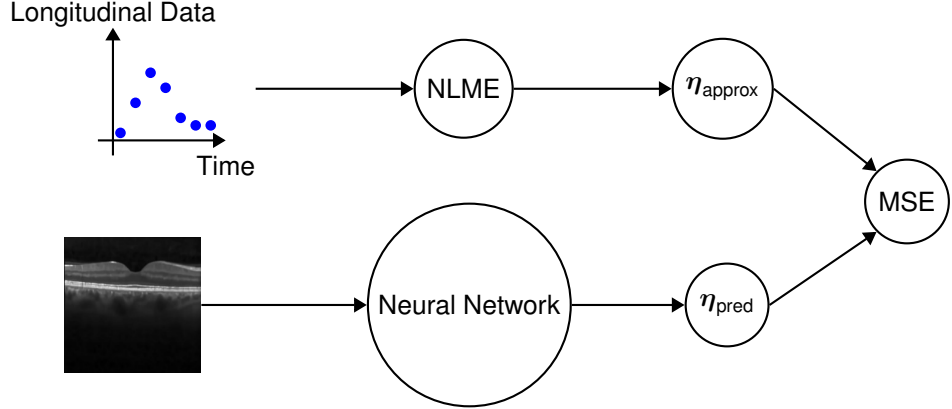}};

    \node[draw, circle, right=2.2cm of initialImage] (nn) {Neural Network};
    \draw[->] (initialImage.east) -- (nn.west);

    \node[draw, circle, right=1.5cm of nn] (eta_pred) {\( \boldsymbol{\eta}_{\text{pred}} \)};
    \draw[->] (nn.east) -- (eta_pred.west);

    \node[draw, circle, right=1.2cm of eta_approx, yshift=-1.5cm] (mse) {MSE};
    \draw[->] (eta_approx.east) -- (mse.north west);
    \draw[->] (eta_pred.east) -- (mse.south west);

\end{tikzpicture}
\caption{Training procedure leveraging the NLME model. Starting with synthetic longitudinal data, the NLME model approximates the random effects, yielding \( \boldsymbol{\eta_\text{approx}} \). This approximation aids in training the neural network to relate images to their corresponding random effects.}
\label{fig:train_nn}
\end{figure}

For each noise level (0, 1, 9, 18, and 49), we trained a ResNet18 model using \(\boldsymbol{\eta_{\text{approx}}}\) as targets. The process is illustrated in Figure \ref{fig:train_nn}. The model was trained on an  NVIDIA RTX A5500 GPU with a batch size of 16. We used the Adam optimizer with a learning rate of 0.001 and original moment coefficients suggested in the paper \cite{Kingma_Ba_2017}.

\begin{figure}
\centering
\begin{tikzpicture}[auto, thick, >=Triangle, font=\sffamily]

    \node[inner sep=0pt] (initialImage) {\includegraphics[width=1.8cm]{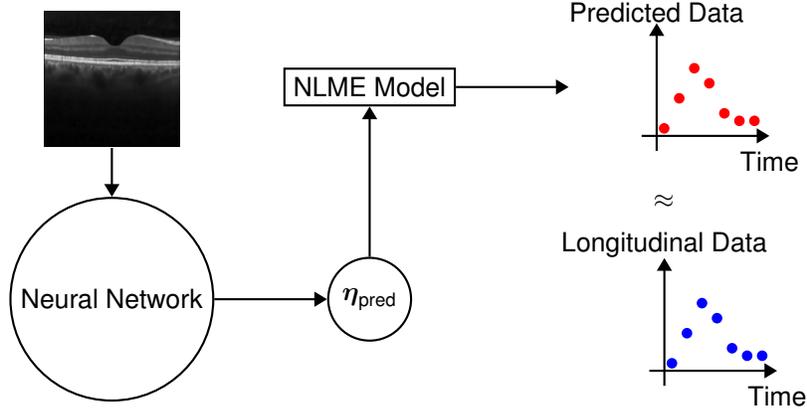}};

    \node[draw, circle, below=0.65cm of initialImage] (nn) {Neural Network};
    \draw[->] (initialImage.south) -- (nn.north);

    \node[draw, circle, right=1.5cm of nn] (eta_pred) {\( \boldsymbol{\eta}_{\text{pred}} \)};
    \draw[->] (nn.east) -- (eta_pred.west);

    \node[draw, rectangle, above=2cm of eta_pred] (nlme) {NLME Model};
    \draw[->] (eta_pred.north) -- (nlme.south);

    \node[inner sep=0pt, right=1.5cm of nlme] (predictedData) {
        \begin{tikzpicture}
            \draw[->, thick] (-0.2,0) -- (1.5,0) node[below=0.2cm] {Time};
            \draw[->, thick] (0,-0.2) -- (0,1.5) node[above] {Predicted Data};
            \foreach \x/\y in {0.1/0.1, 0.3/0.5, 0.5/0.9, 0.7/0.7, 0.9/0.3, 1.1/0.2, 1.3/0.2}{
                \fill[red] (\x,\y) circle (2pt);
            }
        \end{tikzpicture}
    };
    \draw[->] (nlme.east) -- (predictedData.west);

    \node[inner sep=0pt, below=0.8cm of predictedData] (longitudinalData) {
        \begin{tikzpicture}
            \draw[->, thick] (-0.2,0) -- (1.5,0) node[below=0.2cm] {Time};
            \draw[->, thick] (0,-0.2) -- (0,1.5) node[above] {Longitudinal Data};
            \foreach \x/\y in {0.1/0.1, 0.3/0.5, 0.5/0.9, 0.7/0.7, 0.9/0.3, 1.1/0.2, 1.3/0.2}{
                \fill[blue] (\x,\y) circle (2pt);
            }
        \end{tikzpicture}
    };

    \path (longitudinalData.north) -- node[left] {$\approx$} (predictedData.south);

\end{tikzpicture}
\caption{Procedure for predicting longitudinal data using the trained model. An image, when introduced to the trained neural network, predicts the associated random effects, denoted as \( \boldsymbol{\eta}_{\text{pred}} \). These effects, when incorporated into the NLME model, yield the corresponding longitudinal data. The predicted longitudinal data can then be compared with the original synthetic longitudinal data for validation.}
\label{fig:predict_longitudinal}
\end{figure}

After training, we assessed the performance of our models on the test sets corresponding to the five different noise levels. The prediction workflow is depicted in Figure \ref{fig:predict_longitudinal}. For each test image, we used the trained ResNet18 model to predict the random effects. These predicted random effects were then utilized in the NLME model to generate predicted longitudinal data. Finally, we compared the predicted longitudinal data with the original synthetic longitudinal data to evaluate the model's performance across the different noise levels.

\section{Results}
\begin{figure}[htbp]
\centering
\includegraphics[width=1.0\textwidth]{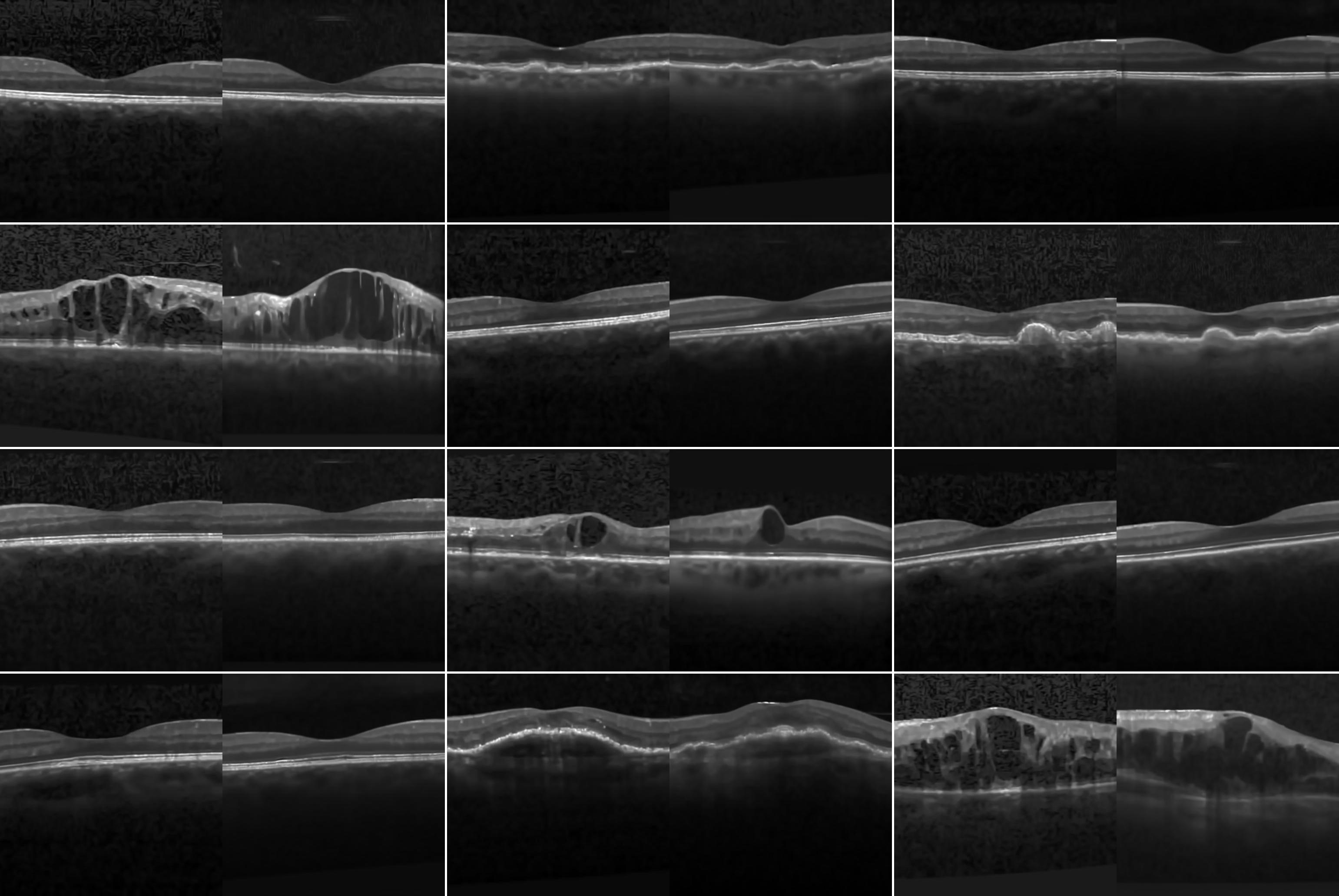}
\caption{Comparison of original OCT images from the validation set (left) and their respective reconstructions (right) derived from the SD model when it was conditioned on the latent space vector $\boldsymbol{z}$ created by the VAE encoder.}
\label{fig:reconstructed_images}
\end{figure}

To validate the efficacy of our model, we first evaluate the SD model's ability to reconstruct original OCT images. This is achieved by encoding real OCT images from the test set using the VAE encoder to generate $\boldsymbol{z}$. These latent representations are then used as conditions in the SD model to produce the reconstructed images. Figure \ref{fig:reconstructed_images} shows original OCT images from the test set side by side with their corresponding reconstructions from the SD model. While an in-depth visual assessment is beyond this study's scope, a preliminary observation indicates that most reconstructed images bear a qualitative resemblance to the originals. Nonetheless, differences and variations can be discerned.

\begin{figure}[htbp]
\centering
\begin{tikzpicture}
    \node[anchor=south west,inner sep=0] (image) at (0,0) {\includegraphics[width=1.0\textwidth]{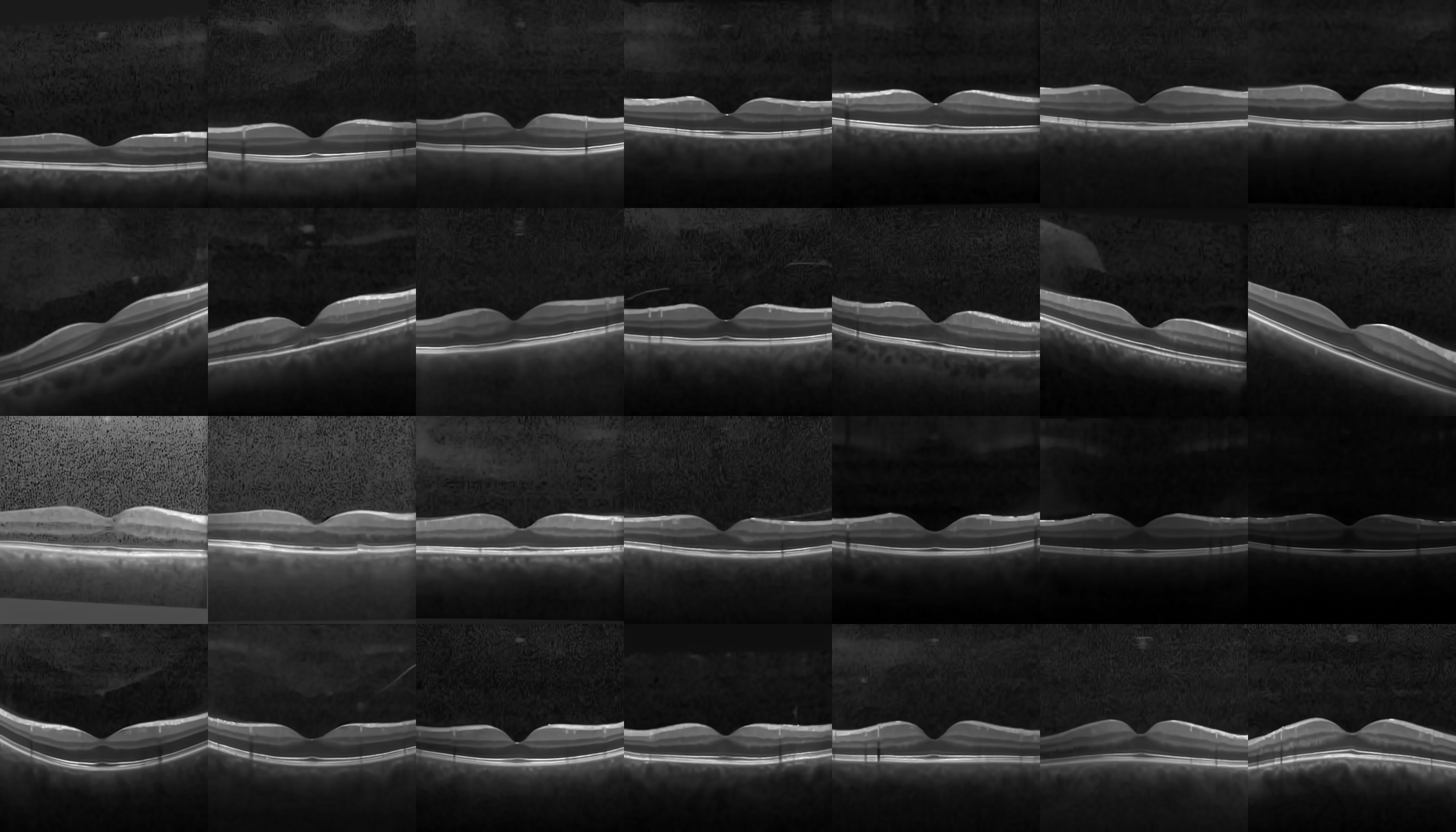}};

    \pgfmathsetmacro{\imagewidth}{14} 
    \pgfmathsetmacro{\imageheight}{10} 

    \node[anchor=north, yshift=-0.25cm] at (\imagewidth/2,0) {$z_i$};

    \draw[->, ultra thick] (\imagewidth/2 - 1, -0.5) -- (1.0, -0.5); 
    \draw[->, ultra thick] (\imagewidth/2 + 1, -0.5) -- (\imagewidth - 1.0, -0.5); 
\end{tikzpicture}
\caption{Generated OCT images delineating the effects of selected latent variable dimensions. Rows, in descending order, depict modifications in vertical positioning, tilt, luminance, and curvature. The central column, with the selected latent variables equal to zero, serves as a reference. Horizontal deviations from the central column signify alterations in a singular latent variable dimension, with all other dimensions held constant.}
\label{fig:altered_images}
\end{figure}

Figure \ref{fig:altered_images} displays a \( 4 \times 7
\) grid of synthetically generated OCT images, demonstrating the effects of varying four distinct latent variable dimensions. The central column of images is generated by taking a latent vector \( \boldsymbol{z} \) drawn from a standard normal distribution \( \mathcal{N}(\boldsymbol{0},\boldsymbol{1}) \), setting the values in the chosen latent dimensions to zero, and then passing this modified vector through the SD model. This central column serves as a baseline for comparison, highlighting the effects of changing individual latent dimensions $z_i$ as one moves horizontally across the grid.

Table \ref{tab:SD_as_decoder} demonstrates the performance in predicting the random effects under different noise levels. As expected, the $R^2$ value decreases with increased noise. However, the fraction of the theoretical maximal explainability decreases at a much slower rate, indicating that the model can still extract a significant portion of the signal even when only 2\% of the signal is present in the data.

\begin{table}[htbp]
\caption{Comparison of the predicted random effects to the true random effects under varying noise levels. The $R^2$ decreases with increased noise, but the fraction of theoretical maximal explainability decreases at a slower rate, indicating that the model can still extract a significant portion of the signal even at high noise levels. The 95\% confidence intervals were obtained by bootstrapping the test set 1000 times.}
\label{tab:SD_as_decoder}
\centering
\begin{tabular}{ccccccc}
\toprule
\multirow{2}{*}{\(\sigma^2\)} & \multicolumn{2}{c}{Comparison} &  &  & Theoretical & Fraction \\
& Term 1 & Term 2 & MSE & $R^2$ & Max \( \frac{1}{1+\sigma^2} \) & of Max \\
\midrule
\multirow{3}{*}{0}  & \(\boldsymbol{\hat{\eta}}\) & \(\boldsymbol{\eta_{\text{pred}}}\) & 0.3993 \textcolor{gray}{\scriptsize[0.3973, 0.4014]} & 0.6007 \textcolor{gray}{\scriptsize[0.5990, 0.6025]} & 1.0000 & 0.6007 \\
                    & \(\boldsymbol{\hat{\eta}}\) & \(\boldsymbol{\eta_{\text{approx}}}\) & 0.3048 \textcolor{gray}{\scriptsize[0.3023, 0.3075]} & 0.6952 \textcolor{gray}{\scriptsize[0.6927, 0.6975]} & & \\
                    & \(\boldsymbol{\eta_{\text{approx}}}\) & \(\boldsymbol{\eta_{\text{pred}}}\) & 0.2383 \textcolor{gray}{\scriptsize[0.2370, 0.2397]} & 0.6465 \textcolor{gray}{\scriptsize[0.6444, 0.6488]} & & \\
\midrule
\multirow{3}{*}{1}  & \(\boldsymbol{\hat{\eta}}\) & \(\boldsymbol{\eta_{\text{pred}}}\) &  0.6983 \textcolor{gray}{\scriptsize[0.6948, 0.7018]} & 0.3017  \textcolor{gray}{\scriptsize[0.2993, 0.3038]}  & 0.5000 & 0.6035 \\
                    & \(\boldsymbol{\hat{\eta}}\) & \(\boldsymbol{\eta_{\text{approx}}}\) &  0.3044 \textcolor{gray}{\scriptsize[0.3020, 0.3069]} & 0.6956 \textcolor{gray}{\scriptsize[0.6932, 0.6980]}  & & \\
                    & \(\boldsymbol{\eta_{\text{approx}}}\) & \(\boldsymbol{\eta_{\text{pred}}}\) & 0.4691 \textcolor{gray}{\scriptsize[0.4667, 0.4716]} & 0.3042 \textcolor{gray}{\scriptsize[0.3012, 0.3072]} & & \\
\midrule
\multirow{3}{*}{9}  & \(\boldsymbol{\hat{\eta}}\) & \(\boldsymbol{\eta_{\text{pred}}}\) & 0.9456 \textcolor{gray}{\scriptsize[0.9409, 0.9503]} & 0.0544 \textcolor{gray}{\scriptsize[0.0530, 0.0557]}  & 0.1000 & 0.5436 \\
                    & \(\boldsymbol{\hat{\eta}}\) & \(\boldsymbol{\eta_{\text{approx}}}\) & 0.3052 \textcolor{gray}{\scriptsize[0.3027, 0.3075]} & 0.6948 \textcolor{gray}{\scriptsize[0.6927, 0.6972]}  & & \\
                    & \(\boldsymbol{\eta_{\text{approx}}}\) & \(\boldsymbol{\eta_{\text{pred}}}\) & 0.5720 \textcolor{gray}{\scriptsize[0.5692, 0.5750]} & 0.0599 \textcolor{gray}{\scriptsize[0.0580, 0.0617]}  & & \\
\midrule
\multirow{3}{*}{18} & \(\boldsymbol{\hat{\eta}}\) & \(\boldsymbol{\eta_{\text{pred}}}\) & 0.9729 \textcolor{gray}{\scriptsize[0.9680, 0.9775]} & 0.0271 \textcolor{gray}{\scriptsize[0.0261, 0.0281]}  & 0.0526 & 0.5149 \\
                    & \(\boldsymbol{\hat{\eta}}\) & \(\boldsymbol{\eta_{\text{approx}}}\) &  0.3035 \textcolor{gray}{\scriptsize[0.3009, 0.3060]} & 0.6965  \textcolor{gray}{\scriptsize[0.6942, 0.6989]}  & &  \\
                    & \(\boldsymbol{\eta_{\text{approx}}}\) & \(\boldsymbol{\eta_{\text{pred}}}\) &  0.6520 \textcolor{gray}{\scriptsize[0.6489, 0.6550]} & 0.0312 \textcolor{gray}{\scriptsize[0.0301, 0.0325]} & & \\
\midrule
\multirow{3}{*}{49} & \(\boldsymbol{\hat{\eta}}\) & \(\boldsymbol{\eta_{\text{pred}}}\) &  0.9936 \textcolor{gray}{\scriptsize[0.9888, 0.9985]} & 0.0064 \textcolor{gray}{\scriptsize[0.0056, 0.0072)]}  &  0.0200 & 0.3201 \\
                    & \(\boldsymbol{\hat{\eta}}\) & \(\boldsymbol{\eta_{\text{approx}}}\) & 0.3049 \textcolor{gray}{\scriptsize[0.3022, 0.3077]} & 0.6951 \textcolor{gray}{\scriptsize[0.6927, 0.6975]} & &\\
                    & \(\boldsymbol{\eta_{\text{approx}}}\) & \(\boldsymbol{\eta_{\text{pred}}}\) &  0.6657 \textcolor{gray}{\scriptsize[0.6623, 0.6688]} & 0.0124  \textcolor{gray}{\scriptsize[0.0116, 0.0132]}  & & \\
\bottomrule
\end{tabular}
\end{table}

To further validate the robustness of our results, we employ the NLME model to generate longitudinal data using the predicted random effects, as shown in Table \ref{tab:nll_results}. We evaluate the performance using the conditional negative log-likelihood (NLL), conditioning on the true data without noise. We compare the NLL of the true data with noise, the approximated data, the average data (all random effects set to 0), and random data (all random effects drawn from a standard normal distribution). The results show that even for the highest noise level (2\% signal), the NLL of the predicted data is lower than that of the average data, demonstrating that the predicted random effects effectively improve the prediction of longitudinal data. It is worth noting that as noise is added, the NLL of the true and approximated data approach that of the random data. This is because they maintain their variance close to unity, while the predicted data, for example, have a variance of 0.00147 for the maximum noise level.

\begin{table}[h]
\centering
\caption{Comparison of conditional NLL of longitudinal data generated from predicted random effects versus true data with noise, approximated data, average data (all random effects set to 0), and random data (all random effects drawn from a standard normal distribution). The predicted random effects result in lower NLL than the average data even at high noise levels, demonstrating their effectiveness in improving longitudinal predictions. The 95\% confidence intervals for the NLL values were obtained by bootstrapping the test set 1000 times.}
\footnotesize
\begin{tabular}{cccccc}
\toprule
Noise Level & True & Predicted & Approximate & Average & Random \\ \midrule
0 & -67 \textcolor{gray}{\scriptsize[-67, -67]} & -59 \textcolor{gray}{\scriptsize[-60, -59]} & -63 \textcolor{gray}{\scriptsize[-63, -63]} & 1616 \textcolor{gray}{\scriptsize[1587, 1647]} & 2896 \textcolor{gray}{\scriptsize[2861, 2934]} \\
1 & 1178 \textcolor{gray}{\scriptsize[1159, 1200]} & 372 \textcolor{gray}{\scriptsize[363, 380]} & 1159 \textcolor{gray}{\scriptsize[1140, 1178]} & 1616 \textcolor{gray}{\scriptsize[1587, 1647]} & 2915 \textcolor{gray}{\scriptsize[2874, 2954]} \\
9 & 2299 \textcolor{gray}{\scriptsize[2270, 2333]} & 1170 \textcolor{gray}{\scriptsize[1145, 1194]} & 2249 \textcolor{gray}{\scriptsize[2214, 2282]} & 1616 \textcolor{gray}{\scriptsize[1586, 1645]} & 2899 \textcolor{gray}{\scriptsize[2863, 2935]} \\
18 & 2478 \textcolor{gray}{\scriptsize[2444, 2510]} & 1287 \textcolor{gray}{\scriptsize[1260, 1313]} & 2430 \textcolor{gray}{\scriptsize[2395, 2466]} & 1616 \textcolor{gray}{\scriptsize[1586, 1647]} & 2926 \textcolor{gray}{\scriptsize[2886, 2968]} \\
49 & 2622 \textcolor{gray}{\scriptsize[2586, 2658]} & 1410 \textcolor{gray}{\scriptsize[1381, 1438]} & 2584 \textcolor{gray}{\scriptsize[2546, 2620]} & 1616 \textcolor{gray}{\scriptsize[1583, 1647]} & 2906 \textcolor{gray}{\scriptsize[2868, 2946]} \\ \bottomrule
\end{tabular}
\label{tab:nll_results}
\end{table}

\section{Conclusion}

Our framework enables testing and benchmarking of predictive models in a controlled setting, addressing data privacy concerns by eliminating the need for real patient data during initial research and model development phases. This approach serves as a valuable tool for developing and refining predictive models in healthcare, providing a structured method to understand the impact of complex covariates on longitudinal patient observations.

\newpage
\bibliographystyle{unsrt}
\bibliography{references}

\begin{thebibliography}{10}

\bibitem{LeCun_Bengio_Hinton_2015}
Yann LeCun, Yoshua Bengio, and Geoffrey Hinton.
\newblock Deep learning.
\newblock {\em Nature}, 521(7553):436–444, May 2015.

\bibitem{Deng_Dong_Socher_ImageNet}
Jia Deng, Wei Dong, Richard Socher, Li-Jia Li, Kai Li, and Li~Fei-Fei.
\newblock Imagenet: A large-scale hierarchical image database.
\newblock In {\em 2009 IEEE Conference on Computer Vision and Pattern
  Recognition}, page 248–255, Jun 2009.

\bibitem{Lin_Maire_Belongie_Hays_Perona_Ramanan_Dollar_Zitnick_2014}
Tsung-Yi Lin, Michael Maire, Serge Belongie, James Hays, Pietro Perona, Deva
  Ramanan, Piotr Dollár, and C.~Lawrence Zitnick.
\newblock Microsoft coco: Common objects in context.
\newblock In David Fleet, Tomas Pajdla, Bernt Schiele, and Tinne Tuytelaars,
  editors, {\em Computer Vision – ECCV 2014}, page 740–755, Cham, 2014.
  Springer International Publishing.

\bibitem{LAION_2022}
Christoph Schuhmann, Romain Beaumont, Richard Vencu, Cade Gordon, Ross
  Wightman, Mehdi Cherti, Theo Coombes, Aarush Katta, Clayton Mullis, Mitchell
  Wortsman, Patrick Schramowski, Srivatsa Kundurthy, Katherine Crowson, Ludwig
  Schmidt, Robert Kaczmarczyk, and Jenia Jitsev.
\newblock Laion-5b: An open large-scale dataset for training next generation
  image-text models.
\newblock {\em Advances in Neural Information Processing Systems},
  35:25278–25294, December 2022.

\bibitem{Mesaros_Heittola_Virtanen_2018}
Annamaria Mesaros, Toni Heittola, and Tuomas Virtanen.
\newblock A multi-device dataset for urban acoustic scene classification.
\newblock {\em arXiv preprint arXiv:1807.09840}, 2018.

\bibitem{Rakotomamonjy_Gasso_2015}
Alain Rakotomamonjy and Gilles Gasso.
\newblock Histogram of gradients of time-frequency representations for audio
  scene detection.
\newblock {\em CoRR}, abs/1508.04909, 2015.

\bibitem{Koren_Bell_Volinsky_2009}
Yehuda Koren, Robert Bell, and Chris Volinsky.
\newblock Matrix factorization techniques for recommender systems.
\newblock {\em Computer}, 42(8):30–37, August 2009.

\bibitem{Koenigstein_Dror_Koren_2011}
Noam Koenigstein, Gideon Dror, and Yehuda Koren.
\newblock Yahoo! music recommendations: modeling music ratings with temporal
  dynamics and item taxonomy.
\newblock In {\em Proceedings of the fifth ACM conference on Recommender
  systems}, RecSys ’11, page 165–172, New York, NY, USA, October 2011.
  Association for Computing Machinery.

\bibitem{Johnson_Pollard_2016}
Alistair E.~W. Johnson, Tom~J. Pollard, Lu~Shen, Li-wei~H. Lehman, Mengling
  Feng, Mohammad Ghassemi, Benjamin Moody, Peter Szolovits, Leo Anthony~Celi,
  and Roger~G. Mark.
\newblock Mimic-iii, a freely accessible critical care database.
\newblock {\em Scientific Data}, 3(1):160035, May 2016.

\bibitem{kingma2013auto}
Diederik~P Kingma and Max Welling.
\newblock Auto-encoding variational bayes.
\newblock {\em arXiv preprint arXiv:1312.6114}, 2013.

\bibitem{rombach2022high}
Robin Rombach, Andreas Blattmann, Dominik Lorenz, Patrick Esser, and Bj{\"o}rn
  Ommer.
\newblock High-resolution image synthesis with latent diffusion models.
\newblock In {\em Proceedings of the IEEE/CVF conference on computer vision and
  pattern recognition}, pages 10684--10695, 2022.

\bibitem{lindstrom1990nonlinear}
Mary~J Lindstrom and Douglas~M Bates.
\newblock Nonlinear mixed effects models for repeated measures data.
\newblock {\em Biometrics}, pages 673--687, 1990.

\bibitem{kermany2018large}
Daniel Kermany, Kang Zhang, and Michael Goldbaum.
\newblock Large dataset of labeled optical coherence tomography (oct) and chest
  x-ray images.
\newblock {\em Mendeley Data}, 3(10.17632), 2018.

\bibitem{Higgins_Matthey_2016}
Irina Higgins, Loic Matthey, Arka Pal, Christopher Burgess, Xavier Glorot,
  Matthew Botvinick, Shakir Mohamed, and Alexander Lerchner.
\newblock beta-vae: Learning basic visual concepts with a constrained
  variational framework.
\newblock In {\em International conference on learning representations}, Nov
  2016.

\bibitem{Pihlgren_Nikolaidou_Chhipa_Abid_Saini_Sandin_Liwicki_2023}
Gustav~Grund Pihlgren, Konstantina Nikolaidou, Prakash~Chandra Chhipa, Nosheen
  Abid, Rajkumar Saini, Fredrik Sandin, and Marcus Liwicki.
\newblock A systematic performance analysis of deep perceptual loss networks:
  Breaking transfer learning conventions.
\newblock {\em arXiv preprint arXiv:2302.04032}, 2023.

\bibitem{vgg16}
Karen Simonyan and Andrew Zisserman.
\newblock Very deep convolutional networks for large-scale image recognition.
\newblock {\em arXiv preprint arXiv:1409.1556}, 2014.

\bibitem{Kingma_Ba_2017}
Diederik~P Kingma.
\newblock Adam: A method for stochastic optimization.
\newblock {\em arXiv preprint arXiv:1412.6980}, 2014.

\bibitem{Burgess_Higgins_2018}
Christopher~P Burgess, Irina Higgins, Arka Pal, Loic Matthey, Nick Watters,
  Guillaume Desjardins, and Alexander Lerchner.
\newblock Understanding disentangling in beta-vae.
\newblock {\em arXiv preprint arXiv:1804.03599}, 2018.

\bibitem{Loshchilov_Hutter_2019}
I~Loshchilov.
\newblock Decoupled weight decay regularization.
\newblock {\em arXiv preprint arXiv:1711.05101}, 2017.

\bibitem{pinheiro2006mixed}
Jos{\'e} Pinheiro and Douglas Bates.
\newblock {\em Mixed-effects models in S and S-PLUS}.
\newblock Springer science \& business media, 2006.

\bibitem{Davidian_Giltinan_2003}
Marie Davidian and David~M. Giltinan.
\newblock Nonlinear models for repeated measurement data: An overview and
  update.
\newblock {\em Journal of Agricultural, Biological, and Environmental
  Statistics}, 8(4):387, Dec 2003.

\bibitem{Demidenko_2013}
Eugene Demidenko.
\newblock {\em Mixed Models: Theory and Applications with R}.
\newblock John Wiley \& Sons, Aug 2013.

\bibitem{Pinheiro_Bates_1995}
José~C. Pinheiro and Douglas~M. Bates.
\newblock Approximations to the log-likelihood function in the nonlinear
  mixed-effects model.
\newblock {\em Journal of Computational and Graphical Statistics},
  4(1):12–35, Mar 1995.

\end{thebibliography}
\appendix

\newpage
\begin{center}
\begin{longtable}{cccccc}
\caption{Detailed architecture of the VAE model. The model comprises of a ResNet18 encoder followed by a custom decoder that employs upsampling and convolutional layers. The table presents information for each layer regarding its type, number of input and output channels, output shape, kernel size, stride, and padding. The ``Channels'' column indicates the number of input channels transitioning to output channels for each layer, symbolized as ``input $\rightarrow$ output''. Kernel, stride, and pad columns are only relevant to convolutional layers.}
\label{tab:VAE_architecture} \\

\hline
\textbf{Layer (type)} & \textbf{Channels} & \textbf{Output Shape} & \textbf{Kernel} & \textbf{Stride} & \textbf{Pad} \\
\hline
\endfirsthead

\multicolumn{6}{c}{{\bfseries Table \thetable\ continued from previous page}} \\
\hline
\textbf{Layer (type)} & \textbf{Channels} & \textbf{Output Shape} & \textbf{Kernel} & \textbf{Stride} & \textbf{Pad} \\
\hline
\endhead

\hline \multicolumn{6}{r}{{Continued on next page}} \\
\endfoot

\hline
\endlastfoot

\multicolumn{6}{c}{\textbf{Encoder Layers}} \\
ResNet18 & 1 $\rightarrow$ 512 & (batch size, 512, 1, 1) & - & - & - \\
ReLU & - & (batch size, 512) & - & - & - \\
\hline
\multicolumn{6}{c}{$\boldsymbol{\mu}$ and $\log\boldsymbol{\sigma}^2$ layers} \\
Dense & 512 $\rightarrow$ 128 & (batch size, 128) & - & - & - \\
Dense & 512 $\rightarrow$ 128 & (batch size, 128) & - & - & - \\
\hline
\multicolumn{6}{c}{\textbf{Decoder Layers}} \\
Dense + ReLU & 128 $\rightarrow$ 50176 & (batch size, 50176) & - & - & - \\
Reshape & - & (batch size, 1024, 7, 7) & - & - & - \\
Conv & 1024 $\rightarrow$ 512 & (batch size, 512, 7, 7) & $3\times 3$ & 1 & 1 \\
BatchNorm & - & (batch size, 512, 7, 7) & - & - & - \\
ReLU & - & - & - & - & - \\
Upsample & - & (batch size, 512, 14, 14) & - & - & - \\
Conv & 512 $\rightarrow$ 512 & (batch size, 512, 14, 14) & $3\times 3$ & 1 & 1 \\
BatchNorm & - & (batch size, 512, 14, 14) & - & - & - \\
ReLU & - & - & - & - & - \\
Conv & 512 $\rightarrow$ 256 & (batch size, 256, 14, 14) & $3\times 3$ & 1 & 1 \\
BatchNorm & - & (batch size, 256, 14, 14) & - & - & - \\
ReLU & - & - & - & - & - \\
Upsample & - & (batch size, 256, 28, 28) & - & - & - \\
Conv & 256 $\rightarrow$ 256 & (batch size, 256, 28, 28) & $3\times 3$ & 1 & 1 \\
BatchNorm & - & (batch size, 256, 28, 28) & - & - & - \\
ReLU & - & - & - & - & - \\
Conv & 256 $\rightarrow$ 128 & (batch size, 128, 28, 28) & $3\times 3$ & 1 & 1 \\
BatchNorm & - & (batch size, 128, 28, 28) & - & - & - \\
ReLU & - & - & - & - & - \\
Upsample & - & (batch size, 128, 56, 56) & - & - & - \\
Conv & 128 $\rightarrow$ 128 & (batch size, 128, 56, 56) & $3\times 3$ & 1 & 1 \\
BatchNorm & - & (batch size, 128, 56, 56) & - & - & - \\
ReLU & - & - & - & - & - \\
Conv & 128 $\rightarrow$ 64 & (batch size, 64, 56, 56) & $3\times 3$ & 1 & 1 \\
BatchNorm & - & (batch size, 64, 56, 56) & - & - & - \\
ReLU & - & - & - & - & - \\
Upsample & - & (batch size, 64, 112, 112) & - & - & - \\
Conv & 64 $\rightarrow$ 64 & (batch size, 64, 112, 112) & $3\times 3$ & 1 & 1 \\
BatchNorm & - & (batch size, 64, 112, 112) & - & - & - \\
ReLU & - & - & - & - & - \\
Conv & 64 $\rightarrow$ 32 & (batch size, 32, 112, 112) & $3\times 3$ & 1 & 1 \\
BatchNorm & - & (batch size, 32, 112, 112) & - & - & - \\
ReLU & - & - & - & - & - \\
Upsample & - & (batch size, 32, 224, 224) & - & - & - \\
Conv & 32 $\rightarrow$ 32 & (batch size, 32, 224, 224) & $3\times 3$ & 1 & 1 \\
BatchNorm & - & (batch size, 32, 224, 224) & - & - & - \\
ReLU & - & - & - & - & - \\
Conv & 32 $\rightarrow$ 1 & (batch size, 1, 224, 224) & $3\times 3$ & 1 & 1 \\
Sigmoid & - & - & - & - & - \\
\end{longtable}
\end{center}

\end{document}